\documentclass[runningheads]{llncs}
\usepackage{xcolor}
\usepackage{xspace}

\newcommand{\keiX}{``\includegraphics[width=0.8em]{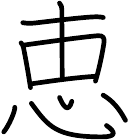}''\xspace}
\newcommand{\ho}{``\includegraphics[width=0.8em]{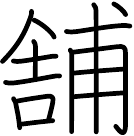}''\xspace}
\newcommand{\hoX}{``\includegraphics[width=0.8em]{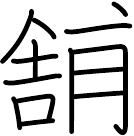}''\xspace}
\newcommand{\tama}{``\includegraphics[width=0.8em]{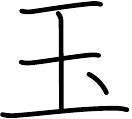}''\xspace}
\newcommand{\king}{``\includegraphics[width=0.8em]{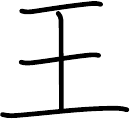}''\xspace}
\newcommand{\kei}{``\includegraphics[width=0.8em]{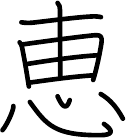}''\xspace}
\newcommand{\charSize}{0.9em}
\newcommand{\charSizeSmall}{0.75em}
\newcommand{\shaX}{``\includegraphics[height=\charSizeSmall]{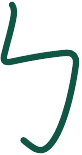}''\xspace}

\newcommand{\shitsuX}{``\includegraphics[height=\charSizeSmall]{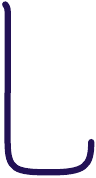}''\xspace}

\newcommand{\tsukuriKATSU}{``\includegraphics[height=\charSizeSmall]{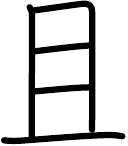}''\xspace}
\newcommand{\henKOME}{``\includegraphics[height=\charSize]{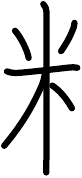}''\xspace}
\newcommand{\henITO}{``\includegraphics[height=\charSize]{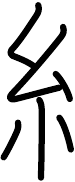}''\xspace}

\newcommand{\arai}{``\includegraphics[height=\charSize]{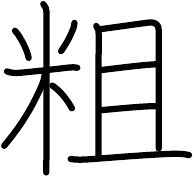}''\xspace}
\newcommand{\kumi}{``\includegraphics[height=\charSize]{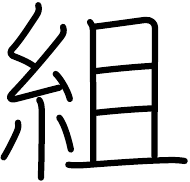}''\xspace}

\usepackage[T1]{fontenc}
\newcommand{\bx}{{\mathbf{x}}}
\usepackage{graphicx}
\begin{document}
\title{Computer-Aided Multi-Stroke Character Simplification by Stroke Removal}
%
%
\author{
Ryo Ishiyama\orcidID{0009-0007-0162-9950} 
\and
Shinnosuke Matsuo\orcidID{0009-0007-2393-186X}
\and
Seiichi Uchida\orcidID{0000-0001-8592-7566}
}
%
\authorrunning{R. Ishiyama et al.}
%
\institute{Kyushu University, Fukuoka, Japan\\
\email{ryo.ishiyama@human.ait.kyushu-u.ac.jp}}
%
\maketitle    
\begin{abstract}
Multi-stroke characters in scripts such as Chinese and Japanese can be highly complex, posing significant challenges for both native speakers and, especially, non-native learners. If these characters can be simplified without degrading their legibility, it could reduce learning barriers for non-native speakers, facilitate simpler and legible font designs, and contribute to efficient character-based communication systems. In this paper, we propose a framework to systematically simplify multi-stroke characters by selectively removing strokes while preserving their overall legibility. More specifically, we use a highly accurate character recognition model to assess legibility and remove those strokes that minimally impact it. 
Experimental results on 1,256 character classes with 5, 10, 15, and 20 strokes reveal several key findings, including the observation that even after removing multiple strokes, many characters remain distinguishable. These findings suggest the potential for more formalized simplification strategies.

\end{abstract}
%
%
\section{Introduction\label{sec:intro}}
\begin{figure}[t]
\centering
\includegraphics[width=0.9\textwidth]{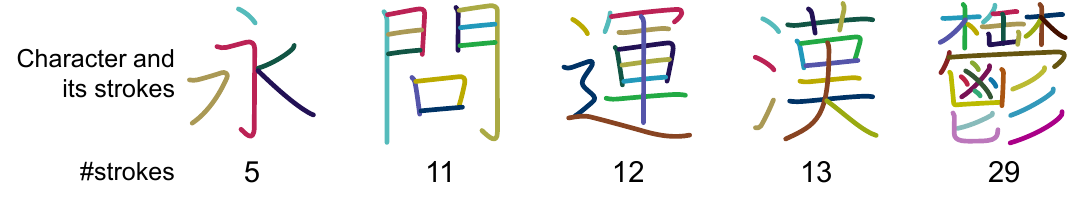}\\[-2mm]
\caption{Examples of multi-stroke characters. Different colors indicate different strokes.}
\label{fig:multi_stroke_examples}
\end{figure}

\begin{figure}[t]
\centering
\includegraphics[width=0.9\textwidth]{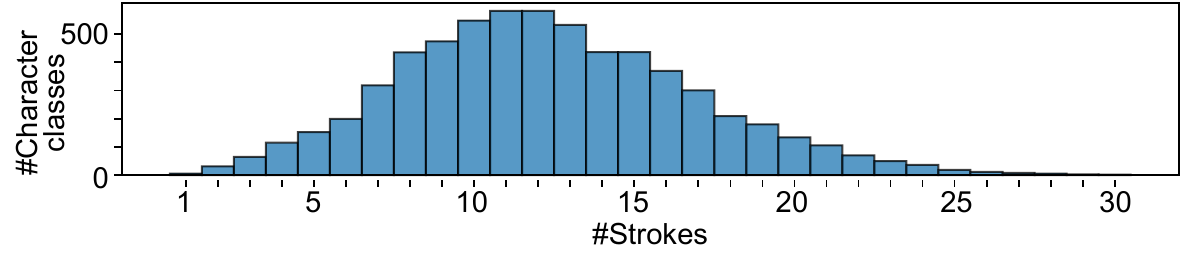}\\[-4mm]
\caption{Distribution of stroke counts.}
\label{fig:stroke_distribution}
\end{figure}

Multi-stroke characters are used in several countries, such as China and Japan.
Fig.~\ref{fig:multi_stroke_examples} shows examples of Japanese multi-stroke characters 
from KanjiVG\footnote{
\texttt{https://kanjivg.tagaini.net/}\label{foot:kanjivg}}, which is a vector graphics dataset of Japanese script. Each stroke is depicted in a different color.
A stroke may be a straight line, a curve, or a corner.
Strokes often intersect in their middle sections and touch at their endpoints.
Fig.~\ref{fig:stroke_distribution} illustrates the distribution of stroke counts for 6,409 multi-stroke characters used in this paper\footnote{Briefly, they cover most of the Japanese multi-stroke characters. These characters will be described in detail in Section~\ref{sec:dataset}.}.
Some characters comprise only a few strokes, whereas others may include up to 30 strokes.
\par

Due to their complex structures, multi-stroke characters can be extremely challenging for both native and non-native speakers to learn. In China, Taiwan, and Japan, approximately 2,500, 2,000, and 1,000 characters are taught over the six years of elementary school, respectively.
Moreover, in all these countries, an additional 1,000 (often more complex) characters are introduced during the three years of junior high school. For non-native learners, mastering a vast number of multi-stroke characters at once can be especially difficult.
\par

Character simplification via the removal of certain strokes has been widely explored to alleviate these challenges. For instance, China officially transitioned from traditional to simplified characters in the 1950s to reduce complexity\cite{StateCouncil1956}. Other simplified forms of Chinese characters have been historically developed and adopted in Japan; for example, \emph{katakana} emerged approximately 1,200 years ago, and \emph{hiragana} appeared about 1,100 years ago. These simplifications significantly reduced character complexity while preserving the distinctiveness of each character.
\par

\begin{figure}[t]
\centering
\includegraphics[width=0.8\textwidth]{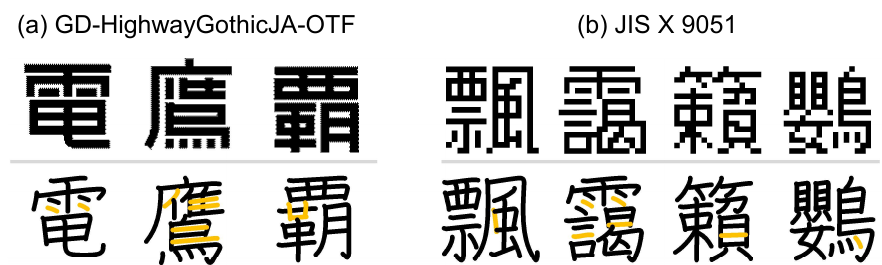}\\[-2mm]
\caption{Fonts featuring character simplification. 
(a)~``Highway Gothic'' for Japanese highway signboards. 
(b)~``JIS X 9051'' designed in a $16\times16$ pixel matrix for low-resolution display.
Below each character is its standard form.
(Orange strokes are removed for simplification.)}
\label{fig:font_with_simplification}
\end{figure}

Simplification is also sometimes applied when developing fonts intended to improve legibility at a distance or on low-resolution displays. For example, the three characters in Fig.~\ref{fig:font_with_simplification}~(a) are taken from a font used on Japanese highway signboards; several strokes were removed to enhance legibility over long distances.
Likewise, the four characters in Fig.~\ref{fig:font_with_simplification}~(b) come from the JIS X 9051 font, in which each character is designed as a $16\times16$ dot matrix for low-resolution screens.
As in (a), several strokes are omitted while preserving the overall structure of each character.
\par

All of these simplifications rely on intuitive or heuristic approaches. To the best of the authors’ knowledge, there has been no systematic attempt at multi-stroke character simplification from a computer science perspective. The Chinese character simplification scheme~\cite{StateCouncil1956} was designed manually (by following several traditions and heuristics). Fonts shown in Figs.~\ref{fig:font_with_simplification}(a) and (b) were manually designed using heuristic insights.
\par

\begin{figure}[t]
\centering
\includegraphics[width=\textwidth]{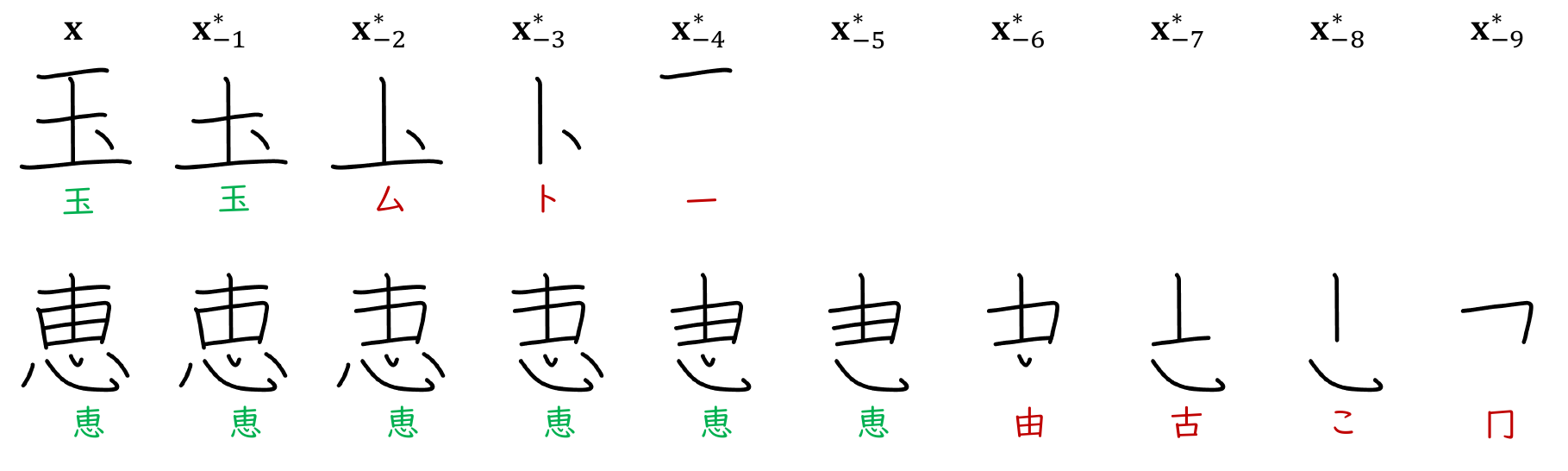}\\[-2mm]
\caption{Examples of multi-stroke character simplification by stroke removal. The leftmost images show the original characters, \tama and \kei. Below each removal result, its classification result is shown. If green, it is still classified as its original class; if red, it is misclassified to a different character class as indicated. Refer to the main text for the meaning of $\bx^*_{-k}$.}
\label{fig:stroke-removal}
\end{figure}

This paper addresses multi-stroke character simplification by \emph{stroke removal}, i.e., omitting certain strokes while maintaining the original legibility. Fig.~\ref{fig:stroke-removal} shows the simplification examples of two multi-stroke characters, \tama (``ball'') and \kei (``grace'').
For the character with fewer strokes (e.g., \tama), removing just one or two strokes significantly affects the whole shape; in contrast, for the character with more strokes (e.g., \kei), removing one stroke does not affect the whole shape. This fact explains why the fonts in Fig.~\ref{fig:font_with_simplification} are acceptable. 
\par

\begin{figure}[t]
\includegraphics[width=\textwidth]{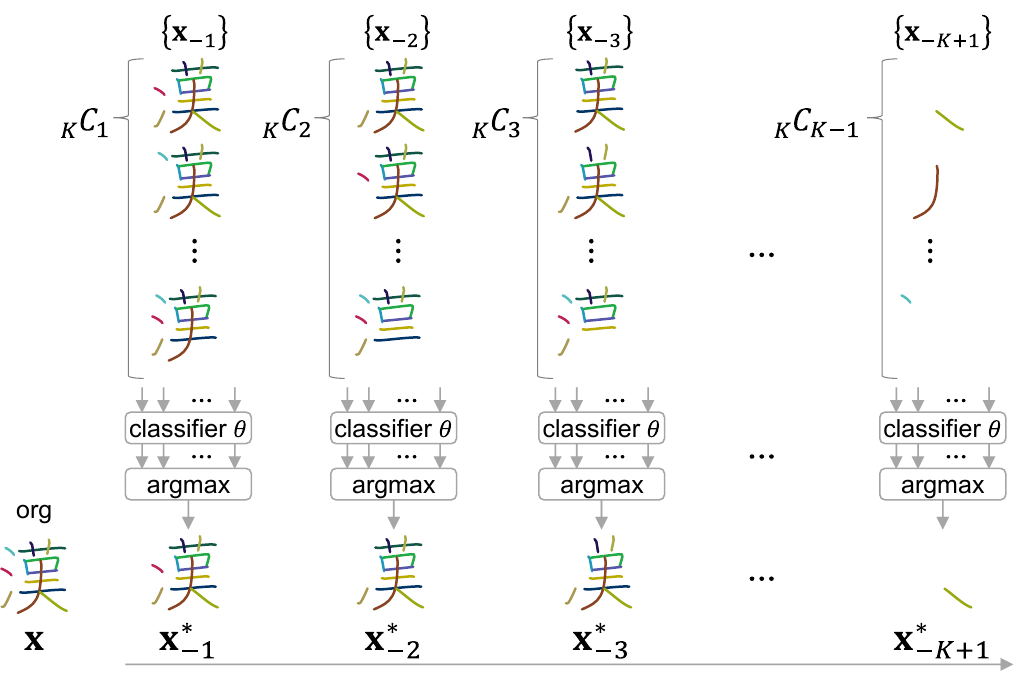}\\[-4mm]
\caption{
Overview of our character simplification approach by stroke removal. The sequence $\bx^*_{-1}, \bx^*_{-2}, \ldots, \bx^*_{-K+1}$ defines the optimal removal sequence.}\label{fig:overview}
\end{figure}

Fig.~\ref{fig:overview} provides an overview of our character simplification approach.
In essence, we consider all possible stroke-removal patterns and evaluate each resulting character’s legibility using a highly accurate multi-stroke classifier, which simulates human perception.
We define the $k$-stroke removal that least compromises legibility as the ``optimal $k$-stroke removal.'' By systematically varying $k$, that is, the number of removed strokes, we can determine, for example, how many strokes can be omitted without significantly degrading legibility.
Further details of this approach are provided in a later section.
\par

In fact, Fig.~\ref{fig:stroke-removal} shows the optimal stroke removal results for \tama and \kei.
The five-stroke character in the upper row, \tama,\ maintains its legibility after one stroke removal and loses it by removing two strokes. It should be noted that \tama has a similar four-stroke character, \king (``king'');  if we first remove the dot-shaped stroke from \tama, it will lose its legibility as \tama and be read as \king. This means the removal order of strokes is important.

Although stroke removal is a simple approach, it offers a clear way to identify which strokes are less critical for legibility (and, conversely, which are crucial). Moreover, if we discover common tendencies across multiple character classes, these findings can be regarded as generalizable rules for character simplification. Such rules could then be applied to produce simpler character sets that are easier for non-native learners to read and write.
\par

Using this approach, we reveal several properties of multi-stroke characters:

\begin{description}
 \item[\textbf{Robustness to stroke removal.}]
    We demonstrate that removing up to about one-third of a character’s total strokes often leaves computed legibility nearly unchanged, indicating that many multi-stroke characters contain a considerable amount of structural redundancy. This also shows the feasibility of producing simplified characters through stroke removal while preserving legibility.

  \item[\textbf{Tolerant vs.\ intolerant characters.}]
    While legibility generally decreases as the number of removed strokes ($k$) grows, the rate of decline differs markedly among character classes. Some are \emph{tolerant} or ``robust,'' retaining high computed legibility even with extensive stroke removal, whereas others are \emph{intolerant} or ``fragile,'' losing legibility after just a few strokes are removed.

  \item[\textbf{Effect of stroke length.}]
    We examine how the lengths (or sizes) of removed strokes affect the resulting legibility. Although shorter strokes tend to be removed first, we find exceptions where even longer strokes are removed early, suggesting that a stroke’s importance depends on its overall contribution to the character’s shape rather than length alone.
\end{description}

\section{Related Work}
\subsection{Heuristic Simplification}

Multi-stroke characters originating in China are used widely across East Asia, including in China, Japan, Korea, Vietnam, and Singapore. Over each country’s history, these characters have undergone various transformations. Han et al.~\cite{Han2022}, for instance, analyze how multi-stroke characters (particularly Chinese characters) evolved over different historical periods.
\par
However, to the best of our knowledge, no attempts have been made to systematically automate multi-stroke character simplification from a computer science perspective. Indeed, most efforts to simplify such characters have relied on heuristic or experience-based designs. Even in recent projects like ``type.\allowbreak normalize.asia''\footnote{\url{https://type.normalize.asia/}}, which aims to unify character shapes by systematically simplifying and abstracting regional differences, many decisions still hinge on subjective or intuitive judgments. 
Teng and Yamada~\cite{Teng2020} investigate character simplification, and Yan et al.~\cite{yan2011} also examine how simplified characters affect reading processes.
Both studies approach the problem from a cognitive psychology perspective rather than a computer science perspective.
This highlights the continuing absence of a fully systematic approach to multi-stroke character simplification.

\subsection{Stroke structural analysis and character recognition}

Within computer science, multi-stroke characters have primarily been studied in the context of recognition. In handwritten text, strokes may merge or split, and prescribed stroke orders are often disregarded, leading to variability in both stroke count and sequence \cite{Cai2006}. Such variations pose a significant challenge for on-line character recognition, prompting the development of robust methods designed to handle them \cite{10.1016/S0031-3203(00)00165-5,Ota2007}. 
\par

Stroke extraction from handwritten images has also been explored; this task is sometimes referred to as {\em stroke recovery} and is considered a difficult inverse problem. Nonetheless, if realized effectively, it could facilitate applications like calligraphy analysis \cite{Li2023StrokeExtraction,WANG2022108416}. Many multi-stroke characters can be viewed as combinations of subcomponents (also known as radicals), typically arranged in a two-dimensional layout \cite{Cao2020ZeroshotHC,Zhang2020RadicalAN}. This composition offers a convenient framework for learning large sets of complex characters. In the realm of computational character recognition, various approaches leverage radical-based information to enhance performance 
\cite{Wang2017RadicalBasedCC,yu2024chinese,zu2022chinese}.
\par

Stroke information has been utilized in machine learning-based font generation methods~\cite{Liu2024DeepCalliFont,xue2025sfgn}.  
For example, StrokeGAN~\cite{zeng2021strokegan} and Diff-Font~\cite{he2024diff} encode stroke information to enhance generation quality.
While traditional fonts are defined by outlines, recent studies have focused on extracting stroke structures and modeling their connectivity for style transformation \cite{10.1145/3505246}. These approaches help bridge the gap between stroke-based character representation and outline-based representation, enabling more adaptable and expressive font stylization.
\par

As noted above, although the stroke structures of multi-stroke characters have been applied in recognition and generation tasks, there appears to be no existing computational research on their simplification. In this paper, we leverage a highly accurate character recognition model to quantitatively and reproducibly measure character legibility, and we explore to what extent characters can be simplified through stroke removal. We believe this first-of-its-kind endeavor will provide valuable insights for future approaches to multi-stroke character simplification.

\section{Character Simplification by Stroke Removal}
\subsection{Computed legibility}
Fig.~\ref{fig:overview} illustrates an overview of our character simplification approach by stroke removal. Let $\bx$ denote a character from class $c_\bx \in [1, C]$ with $K$ strokes and $\bx_{-k}$ denote a simplified character by removing $k \in [1,K-1]$ strokes from $\bx$. We have $_KC_k$ different candidates for $\bx_{k}$. Among them, we choose $\bx^*_{-k}$ with the highest legibility as the original character class $c_\bx$.\par
Here, we use {\em computed legibility} as an approximation of human legibility. Nowadays, we have multi-stroke character classifiers with near-human recognition accuracy by using deep neural network models, and we assume that they automatically evaluate the legibility of individual character images. More formally, we use a pretrained character classifier $\theta$, which gives the posterior probability $P_\theta(c_\bx \mid \bx)$ as the computed legibility of a non-simplified character $\bx$. Note that $P_\theta(c_\bx \mid \bx_{-k})$ in Eq.~(\ref{eq:cl}) is obtained by applying the softmax function to the unnormalized scores (i.e., logits) from the final layer of the classifier model. Since our interest is the legibility of simplified characters $\bx_{-k}$, we evaluate their computed legibility $P_\theta(c_\bx \mid \bx_{-k})$ by using the same classifier $\theta$. \par
\subsection{Optimal $k$-stroke removal and optimal removal sequence}
We expect the simplified character $\bx_{-k}$ to have enough legibility as its original class $c_\bx$. Hereafter, we focus on $\bx^*_{-k}$, which has the highest computed legibility among $_KC_k$ candidates of $\bx_{-k}$, that is, 
\begin{equation}
    \bx^*_{-k} = \mathop{\mathrm{argmax}}_{\bx_{-k}} P_\theta(c_\bx \mid \bx_{-k}).\label{eq:cl}
\end{equation}
The simplified character $\bx^*_{-k}$ corresponds to the {\em optimal $k$-stroke removal} from the original $\bx$. Hereafter, we call $\bx^*_{-1},\ldots, \bx^*_{-k}, \ldots, \bx^*_{-K+1}$ the {\em optimal removal sequence} of the character class $c_\bx$. 
\par

The optimal stroke removal and the optimal removal sequence of a character class $c_\bx$ depend on other character classes. 
As noted in Section~\ref{sec:intro}, the character \tama has a similar character \king. Therefore, as shown in Fig.~\ref{fig:stroke-removal}, the dot-shaped stroke of \tama does not become the candidate for $\bx^*_{-1}$ to avoid confusion with \king.
As clearly shown by this example, the problem of character simplification is not at all a problem that should be dealt with independently for each character.
\par

The optimal removal sequence is not monotonic; that is, the $(K-k-1)$ strokes of $\bx^*_{-k-1}$ are not always a subset of the $(K-k)$ strokes of $\bx^*_{-k}$.
In fact, the two characters in Fig.~\ref{fig:stroke-removal} both exhibit a non-monotonic sequence.
For example, for \tama, $\bx^*_{-4}$ is not a subset of $\bx^*_{-3}$. This indicates that stroke simplification by optimal stroke removal is {\em not} a greedy optimization process in which strokes are removed one by one as $k$ increases\footnote{Interestingly, it is also not guaranteed that the computed legibility $P_\theta(c_\bx \mid \bx_{-k})$ decreases monotonically as $k$ increases.}.

\subsection{Removal tolerance}
As a metric to evaluate how well a character of class $c_\bx$ can maintain its legibility after stroke removal,
we define {\em removal tolerance} as follows:
\begin{equation}
T_{\bx} = \sum_{k\in [1,K-1]} P_\theta(c_\bx \mid \bx^*_{-k}).
\end{equation}
A large $T_{\bx}$ suggests that the computed legibility 
$P_\theta(c_\bx \mid \bx^*_{-k})$ does not decrease quickly as $k$ increases; in other words, the character $\bx$ can maintain its legibility even after more strokes are removed.
In contrast, $T_{\bx} = 0$ indicates that the character completely loses its legibility even if only one (presumably the least harmful) stroke is removed.
%
\section{Experimental Setup}
\subsection{Multi-stroke Character Dataset\label{sec:dataset}}

We use multi-stroke characters whose stroke count distribution is shown in Fig.~\ref{fig:stroke_distribution}.
They are derived from the KanjiVG dataset\footref{foot:kanjivg} because it provides 
stroke-wise information on individual characters.
KanjiVG also contains digits, Latin alphabets, punctuation marks, and the Japanese 
syllabaries (Hiragana and Katakana), so we exclude them by taking the intersection 
of KanjiVG with the CJK unified ideographs (U+4E00--U+9FFF), where CJK stands for Chinese, Japanese, and Korean.
As a result, we obtain 6,409 multi-stroke characters.
Note that these characters do not include simplified Chinese characters, 
which are already simplified and thus not appropriate for our purposes.
\par

In the following experiments, we selected characters with $K=5, 10, 15,$ and $20$ strokes 
from the original set of 6,409. Consequently, we obtained $145, 543, 434,$ and $134$ characters, 
respectively, for a total of 1,256.  We focus on these four representative stroke counts for 
two reasons. The first and most important reason is that we can observe smooth trends as $K$ changes, making it redundant to present results for all possible $K$ values. 
The second and practical reason is that the computational complexity becomes prohibitive when $K$ grows larger. 
Specifically, to have the optimal removal sequence for {\em each} $K$-stroke character, we need $\sum_{k=1}^{K-1} {}_KC_k$ recognition processes, which quickly becomes intractable. (When $K=30$, we need to perform more than one billion recognition processes for each character). Hence, limiting ourselves to $K=5,10,15,20$ not only captures the essential behavior 
of the proposed simplification technique but also keeps the computation feasible.
\par

In KanjiVG, multi-stroke characters are not provided as bitmap images but rather as vector graphics, which can be decomposed into a canonical set of strokes. The characters in Fig.~\ref{fig:multi_stroke_examples} are color-coded by stroke because they are from KanjiVG. Reviewing  Fig.~\ref{fig:multi_stroke_examples}, we can see that some strokes are long or short, straight or curved, and a few even have hooked shapes. Using the data from KanjiVG, each stroke can be handled independently. Therefore, we can generate simplified characters by removing an arbitrary set of strokes. When we compute the computed legibility by inputting these data into the neural network model described below, we convert the vector representation into a $64\times 64$-pixel bitmap image.

\subsection{Character classifier for computed legibility} \label{sec:experimental_setup_framework}

We used a pretrained neural network model $\theta$ for computed legibility. Specifically, we used DaKanji, which is available on GitHub\footnote{\url{https://github.com/CaptainDario/DaKanji-Single-Kanji-Recognition}}. DaKanji takes $64 \times 64$ grayscale input images and classifies them into one of 6,507 classes, covering all characters included in KanjiVG. DaKanji is based on EfficientNet Lite-0. We used the pretrained model as is, without any fine-tuning. We confirmed that DaKanji can recognize character images from KanjiVG with 99.8\% accuracy\footnote{We confirmed DaKanji also achieves around 98\% accuracy for the characters from a different font called ``Hiragino'' --- so, DaKanji does not overfit to KanjiVG.}. This high accuracy indicates that DaKanji is suitable for calculating computed legibility, which is a possible computer-aided approximation of human legibility. \par

We chose DaKanji because recent neural-network-based character recognition models 
are designed for words, texts, or entire images containing multiple texts. 
In contrast, there are very few publicly available models that focus specifically on 
the recognition of single multi-stroke characters. Consequently, DaKanji was practically 
the most suitable pretrained model we could adopt for our purpose.

\section{Experimental Results}
\subsection{Observations of computed legibility}

\begin{figure}[t]
    \centering
    \includegraphics[width=\linewidth]{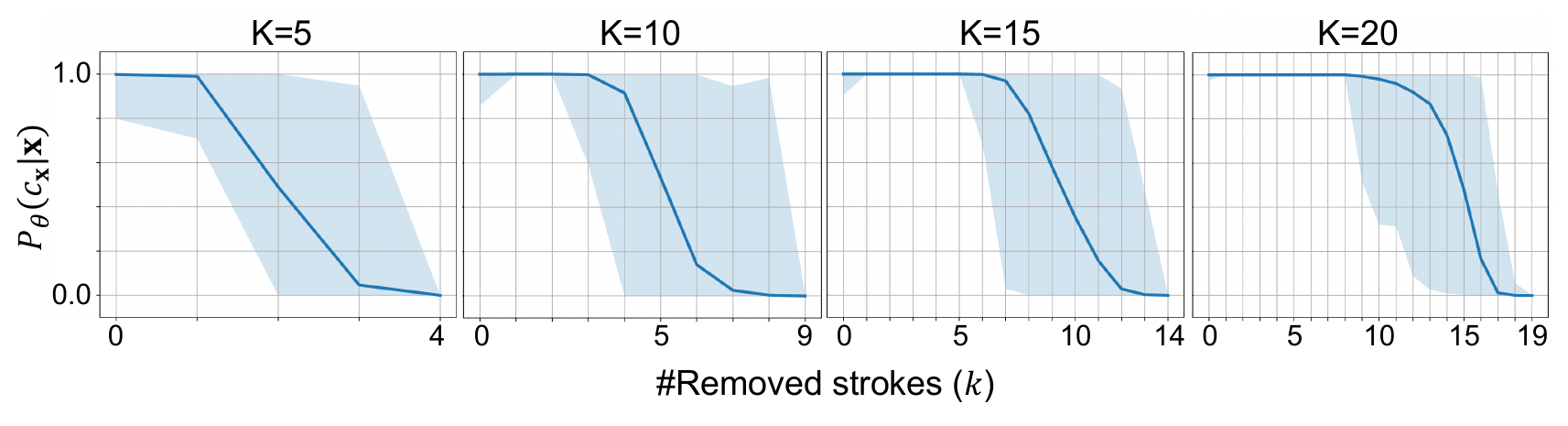}\\[-5mm]
    \caption{
    Computed legibility by the number of removed strokes $k$ for the characters with  $K \in \{5, 10, 15, 20\}$ strokes. \label{fig:legibility_vs_k}}
    \centering  \includegraphics[width=\linewidth]{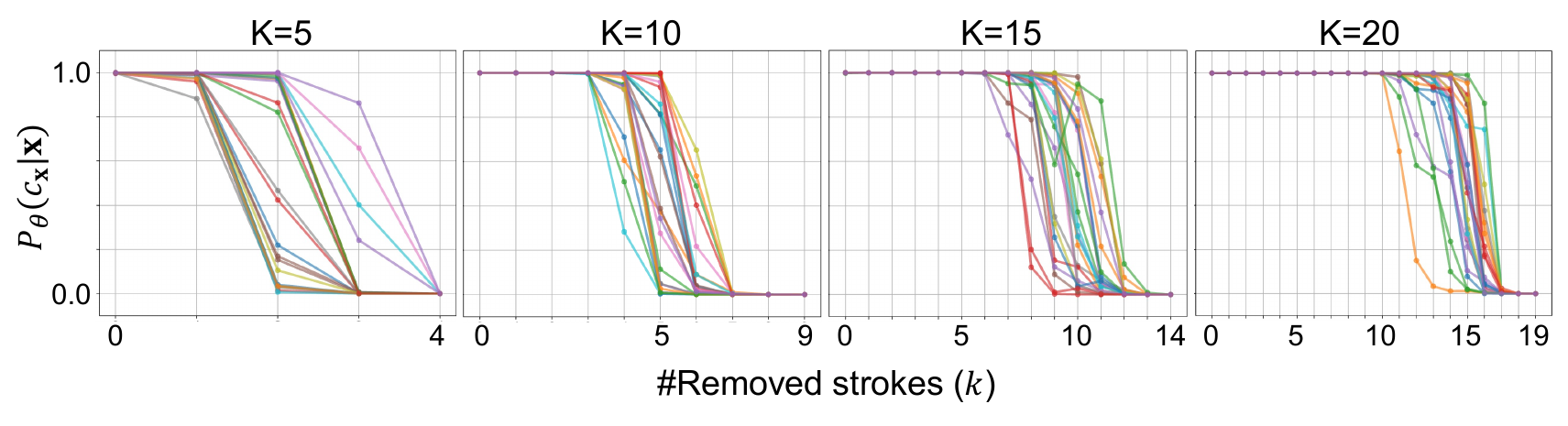}\\[-5mm]
    \caption{Computed legibility for randomly selected 25 characters for each $K \in \{5, 10, 15, 20\}$.}
    \label{fig:parallel}
    \centering
    \includegraphics[width=\linewidth]{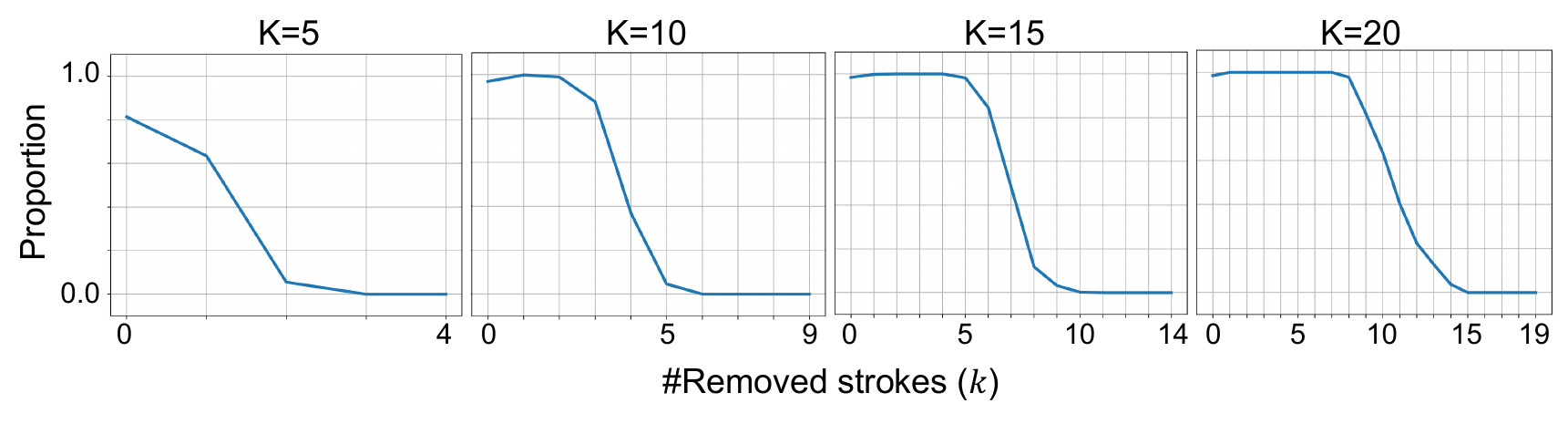}\\[-5mm]
    \caption{
    The proportion of characters whose computed legibility is kept as $1$.}\label{fig:likelihood_1}
\end{figure}

Fig.~\ref{fig:legibility_vs_k} illustrates how the computed legibility changes by the optimal $k$-stroke removal for characters with $K = 5$, $10$, $15$, and $20$ strokes, respectively. In each graph, the bold line indicates the average computed legibility over all characters having $K$ strokes. The light-blue shaded region represents the minimum-to-maximum range of computed legibility for each value of $k$. 
\par

As expected, the computed legibility decreases as $k$ increases. However, when only a few strokes are removed, the computed legibility remains close to $1.0$. This implies that, for many characters, certain strokes exist whose removal does not affect the computed legibility. Notably, for $K=10, 15$ and $20$, the minimum value of computed legibility stays at 1.0 for $k \in \{1,2\}$, $\{1,\dots,5\}$, and $\{1,\dots, 8\}$ respectively, indicating that all characters within those ranges are still recognized correctly after stroke removal without losing its computed legibility at all. 
Interestingly, some characters exhibit higher computed legibility after the removal of one or two strokes than the original character image (i.e., $k = 0$).
For instance, increasing $k$ from 0 to 1 improves the legibility of \ho $\rightarrow$ \hoX from $0.90$ to $1.00$, and of \kei $\rightarrow$ \keiX from $0.86$ to $1.00$.

\par 

This result also shows that characters with a larger number of strokes tend to be more tolerant of stroke removal. One possible explanation is that, analogous to an error-correcting code, such characters have a more redundant structure and can remain legible even if several ``information units'' (i.e., strokes) are missing. In fact, for $K=20$, the original character class can still be correctly recovered from only 8 out of 20 strokes (40\%).
\par

Furthermore, all the average curves exhibit a similar sigmoid-like pattern, regardless of $K$. In other words, there is an initial phase where computed legibility remains unchanged, followed by a rapid decline once two or three strokes are removed, after which the character becomes almost completely unrecognizable. The two or three strokes removed during this steep decline can thus be viewed as \emph{critical} to sustaining the character’s computed legibility.\par
%


\begin{figure}[t]
    \centering  
    \includegraphics[width=\linewidth]{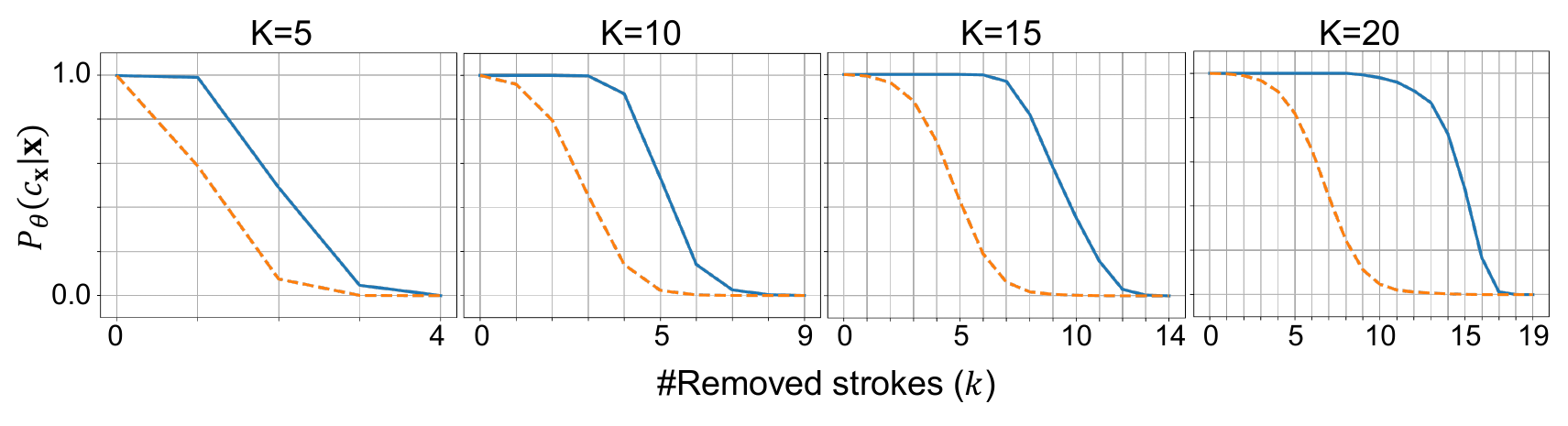}\\[-5mm]
    \caption{Comparison with random $k$-stroke removal. The blue curve is the same as the average curve of Fig.~\ref{fig:legibility_vs_k}. The orange curve is the average computed legibility of random $k$-stroke removal.}
    \label{fig:random-removal}
\end{figure}
Fig.~\ref{fig:parallel} illustrates how the computed legibility of individual characters changes. This visualization further demonstrates that each character’s legibility curve exhibits a sigmoid-like pattern, with certain ``critical strokes'' whose removal dramatically degrades computed legibility.
\par

Fig.~\ref{fig:likelihood_1} shows the proportion of characters whose computed legibility remains exactly $1.0$ even after the optimal $k$-stroke removal. Since this measurement excludes any character whose legibility drops even slightly, it represents a more stringent evaluation than that in Fig.~\ref{fig:legibility_vs_k}. Nevertheless, it is consistent with the overall trends observed in Fig.~\ref{fig:legibility_vs_k}.

\subsection{Comparison with Random $k$-Stroke Removal}
To verify the validity of our optimal $k$-stroke removal strategy in preserving computed legibility, we introduce a baseline method called \emph{random $k$-stroke removal}. As the name implies, this baseline removes $k$ strokes at random from the original set of $K$ strokes. Since there are ${}_K C_k$ ways to choose which $k$ strokes to remove, we compute the computed legibility for \emph{all} such random removals and then take the average. (Conceptually, this corresponds to replacing the argmax operation in Fig.~\ref{fig:overview} with an average.)\par

Fig.~\ref{fig:random-removal} presents the results. The blue curves, as in Fig.~\ref{fig:legibility_vs_k}, represent the average legibility for the optimal $k$-stroke removal, while the orange dashed lines indicate the average legibility under random $k$-stroke removal. In every case ($K = 5, 10, 15,$ and $20$), the computed legibility drops off much more sharply for random removal. As expected, removing strokes indiscriminately leads to a substantial decline in legibility even after removing only a few strokes. For instance, when $K=10$, randomly deleting just three strokes cuts the computed legibility by about half. In contrast, with optimal $k$-stroke removal, the legibility is fully preserved for all tested conditions.

\subsection{Tolerant characters and intolerant characters}
\begin{figure}[t]
    \centering
    \includegraphics[width=\linewidth]{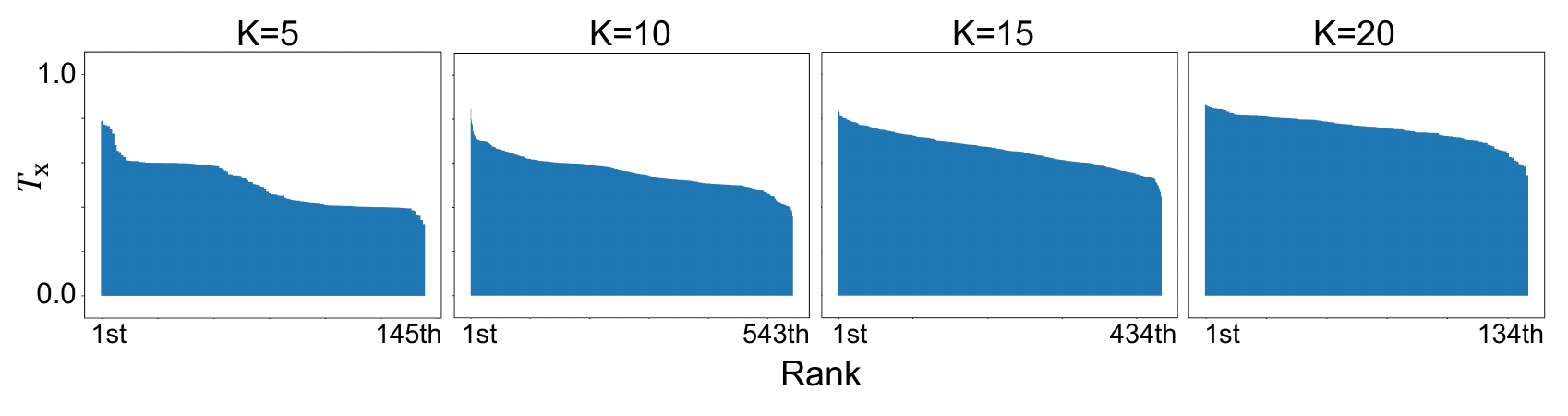}\\[-5mm]
    \caption{Distribution of removal tolerance
    $T_{\bx}$ for all characters with $K\in \{5,10,15,20\}$ strokes. Each plot is a bar chart, where the characters are sorted by $T_{\bx}$ in descending order; therefore, the distribution is monotonically decreasing (more strictly, monotonically non-increasing).
     \label{fig:legibility_vs_k_sum}}
     \vspace{-3mm}
\end{figure}
\begin{figure}[t]
    \centering
    \includegraphics[width=\linewidth]{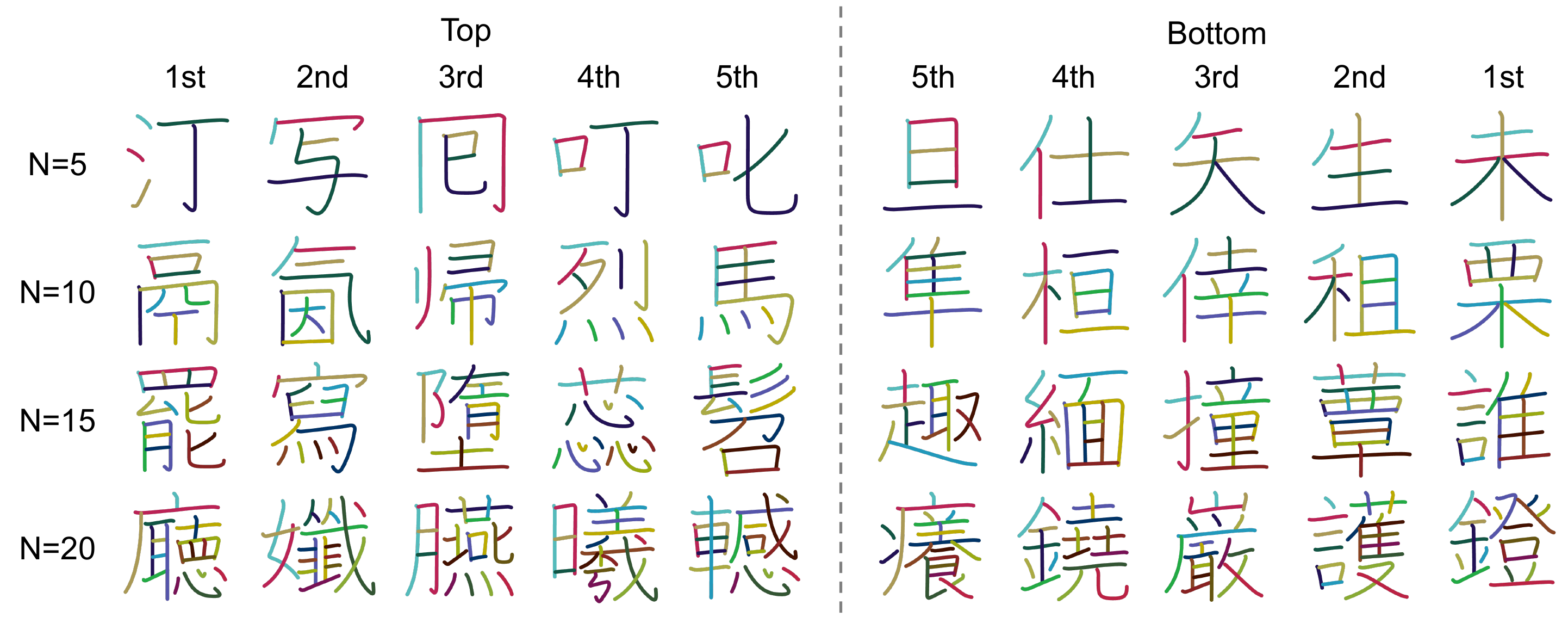}\\[-3mm]
    \caption{Top-5 tolerant and intolerant characters measured by $T_{\bx}$.}
\label{fig:top-five-characters}
\vspace{-3mm}
\end{figure}

Fig.~\ref{fig:legibility_vs_k_sum} shows the distribution of removal tolerance $T_{\bx}$ for all characters with $K\in \{5,10,15,20\}$ strokes. In this plot, the characters are sorted according to $T_{\bx}$ in descending order. Therefore, the leftmost character is the most tolerant to stroke removal, and the rightmost is the least tolerant. The plots prove that $T_{\bx}\sim 0.5$, even for the least tolerant characters. This suggests there is no character with an L-shaped quick drop in its computed legibility --- all characters have a certain tolerance to stroke removals. Several characters have $T_{\bx}\sim 0.8$; it proves that there are characters that could keep their near-original legibility even after many strokes are removed. Finally, $T_\bx$ becomes large according to $K$; this coincides with our previous observation with Fig.~\ref{fig:legibility_vs_k}.
\par

Fig.~\ref{fig:top-five-characters} shows the characters with the top 5 and bottom 5 $T_{\bx}$ values for $N\in \{5, 10, 15, 20\}$.
Examining the top (more tolerant) characters, we often find strokes with a distinctive hook-like structure  (e.g., \shitsuX, \shaX), which may be influenced by the particular font style. In contrast, the 5-stroke characters in the bottom (less tolerant) group tend to include more horizontal strokes than those in the top group. In fact, many multi-stroke characters contain multiple horizontal strokes, and therefore, removing some horizontal strokes makes the original character resemble a different character. Another notable point is that the top characters frequently contain small dot-like strokes that, when omitted, appear to have little impact on overall legibility. Indeed, native writers sometimes skip these minor strokes in casual handwriting.

\subsection{Observations of optimal removal sequences}

\begin{figure}[t]
    \centering
    \includegraphics[width=\linewidth]{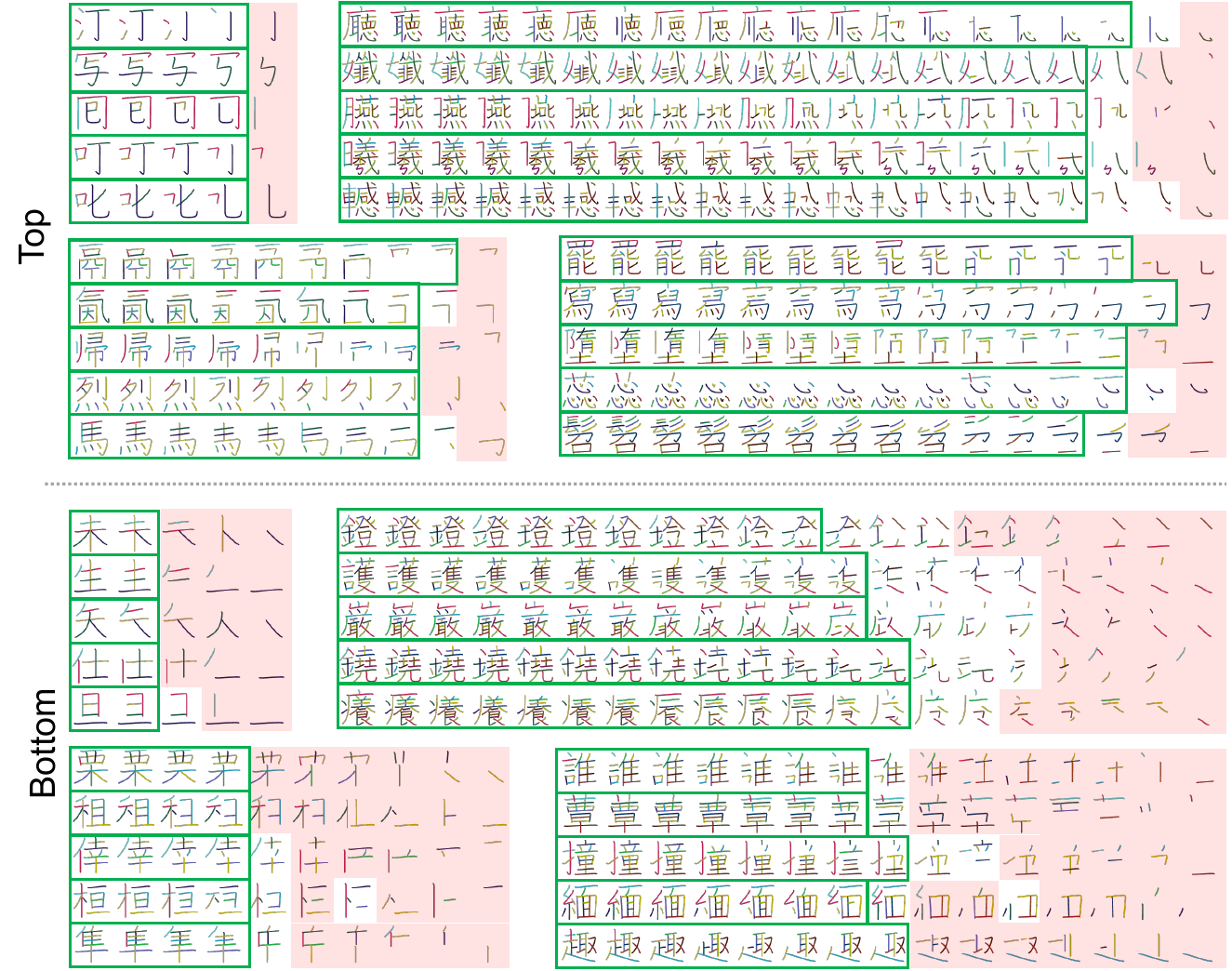}\\[-3mm]
    \caption{The optimal removal sequence for the 40 characters in Fig.~\ref{fig:top-five-characters}. The characters in the green frame are correctly recognized as their original class. The simplified characters with the red background were selected in a situation where all $_KC_k$ candidates had a legibility of less than 1\%. Therefore, the ones shown here were selected (almost accidentally)  with a small margin among candidates with low legibility and better to be ignored when considering their tendency. }
\label{fig:optimal-removal-sequence}
\end{figure}

Fig.~\ref{fig:optimal-removal-sequence} presents the optimal removal sequences for the 40 characters shown in Fig.~\ref{fig:top-five-characters}. Characters with a red background indicate cases where the computed legibility has dropped below 1\%. Since these sequences reflect the ``optimal'' removal orders (i.e., those that maximize computed legibility for each remaining stroke count $K-k$), legibility of only 1\% implies that all other removal patterns (with $K-k$ strokes) would also have legibility of 1\% or less. In other words, if a simplified character has a red background, there might be an alternative shape with only a slight difference in legibility, so focusing on that particular shape carries little significance.
\par

A first observation from Fig.~\ref{fig:optimal-removal-sequence} is that in nearly all examples, a cluster of adjacent strokes---such as those forming a single \emph{radical}---is never removed all at once. For instance, there are no cases where only the top half or the right half of a character is discarded. Instead, strokes tend to be removed incrementally from different parts of the character, thereby preserving its overall shape. One explanation for this phenomenon is that many multi-stroke characters consist of multiple radicals. For example, the character \arai\ is composed of a left radical \henKOME\ and a right radical \tsukuriKATSU, whereas \kumi\ also has \tsukuriKATSU\ on the right but uses a different radical \henITO\ on the left. If the entire left radical were removed first, only the common radical \tsukuriKATSU\ would remain, making it impossible to determine whether the original character was \arai\ or \kumi. Hence, stroke removal tends to proceed in a way that preserves at least part of each radical, echoing the font examples in Fig.~\ref{fig:font_with_simplification}.
\par

A second noteworthy point is that most characters remain correctly recognized as their original class right up until their legibility dips to around 1\%. This ``sudden drop'' phenomenon also appeared in Fig.~\ref{fig:legibility_vs_k}. Even for the bottom group of characters, which are least tolerant of stroke removal, they still tend to be correctly recognized after three (or more) strokes are removed --- except in low-stroke cases (e.g., $K=5$). However, there are instances where native speakers of these multi-stroke characters find the shape unreadable just before it becomes non-legible. This indicates that computed legibility is merely an approximation of human legibility and can, at times, overestimate actual readability.
\par

As noted earlier, the characters with high removal tolerance often contain unique stroke structures (e.g., \shitsuX, \shaX), and Fig.~\ref{fig:optimal-removal-sequence} shows these distinctive strokes frequently remain until the very end---especially pronounced in $K=5$ cases. This suggests that such unique shapes play a key role in sustaining the computed legibility of their original class.

\subsection{Long stroke first? Short stroke first?}
\begin{figure}[t]
    \centering
    \includegraphics[width=\linewidth]{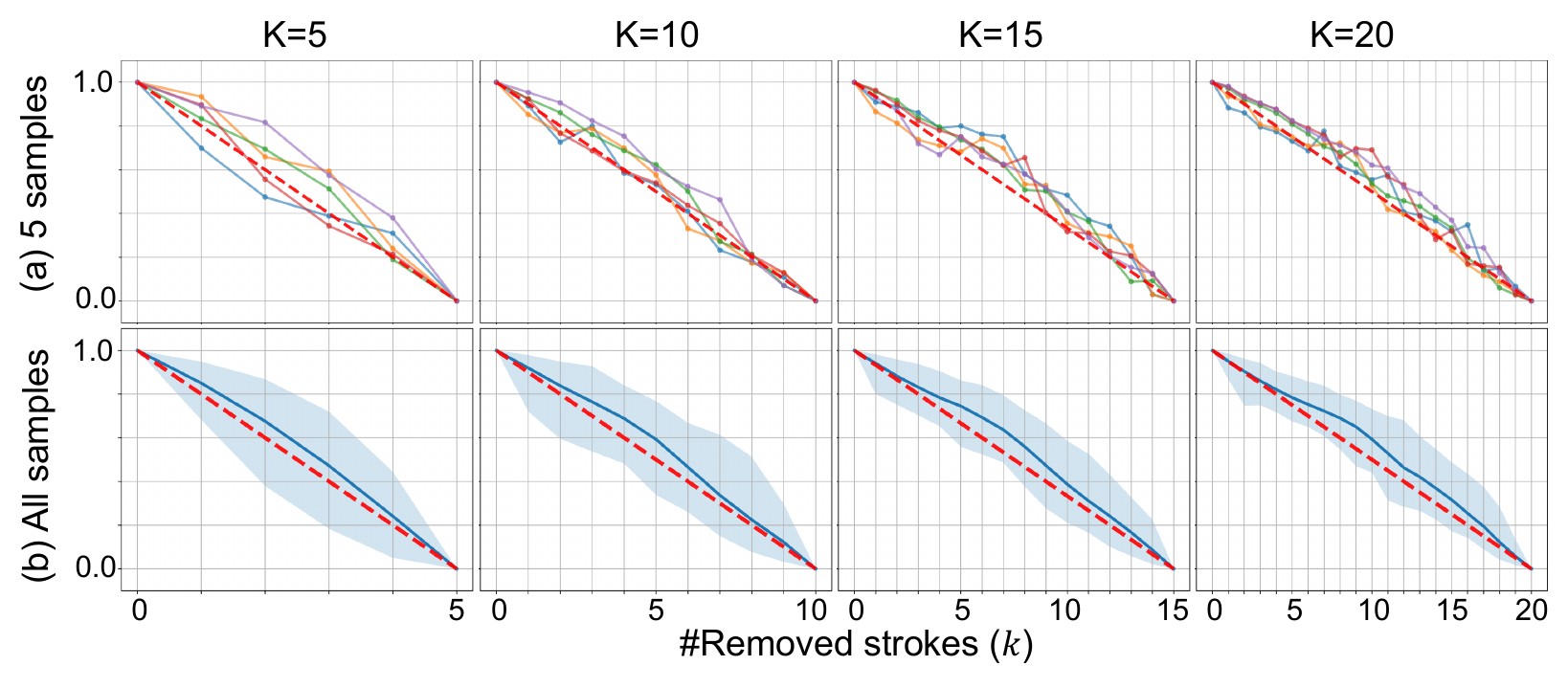}\\[-5mm]
    \caption{
The proportion of stroke (black) pixels remaining in the simplified characters.
(a) Individual plots for five randomly sampled characters.
(b) Plots for all characters with $K$ strokes, where the blue line indicates the average proportion and the shaded region shows the range (minimum to maximum). 
The red dashed line represents $(K - k)/K$, illustrating a linear removal of strokes until they completely disappear.
\label{fig:pixel_removal}}
\end{figure}

Fig.~\ref{fig:pixel_removal} illustrates the proportion of black (stroke) pixels remaining in the simplified characters after each stroke removal. If longer strokes are removed first, the proportion decreases rapidly, creating an ``L-shaped'' curve. Conversely, if shorter strokes are removed first, the curve looks more like ``\rotatebox[origin=c]{180}{L},'' meaning the total pixel count remains high until several short strokes have been eliminated. In (a), the trajectories for five randomly sampled characters vary considerably, indicating no unique removal trends about stroke length. However, the averaged results in (b) (blue line) lie above the red line, suggesting that, on average, shorter strokes tend to be removed before longer ones.

\section{Conclusion, Limitation, and Future Work}

In this paper, we proposed a framework for simplifying multi-stroke characters by stroke removal with a neural network-based character classifier to measure legibility. Specifically, we assumed that the classifier could serve as a proxy for human legibility and analyzed how stroke removal affects recognition. Then, we examined the simplified characters obtained by optimal stroke removal from 1,256 characters with 5, 10, 15, or 20 strokes, and derived the following results:

\begin{itemize}
    \item \textbf{Maintaining legibility up to one-third removal:}
    Even with several strokes removed, legibility often remains high until approximately one-third of the total strokes are gone (e.g., removing two out of five or four out of ten). This phenomenon appears to be related to preserving the character's overall shape.
This behavior resembles the observation in Yan et al.~\cite{yan2011}. 
Their paper on character simplification in cognitive psychology notes that even with 30\% of strokes removed, text remained as readable as the original, 
as long as the overall structure of the character was preserved.

    \item \textbf{Threshold phenomenon:}
    For any given multi-stroke character, there exists a threshold number of removed strokes at which recognition abruptly fails, causing a sudden drop in legibility.

    \item \textbf{Importance of selective removal:}
    A comparison with random $k$-stroke removal demonstrated that indiscriminate deletion significantly reduces legibility after just a few strokes, whereas our optimal removal strategy can maintain near-perfect legibility over the same range of $k$ values.

    \item \textbf{Tolerance related to stroke complexity:}
    Characters with lower tolerance to stroke removal are often composed of simple lines (e.g., straight horizontal or vertical strokes). By contrast, characters with higher tolerance frequently possess more unusual strokes (e.g., \shitsuX, \shaX), which tend to remain until the final stages of removal.

    \item \textbf{Balanced removal preserves global shape:}
    Examining the sequences $x_k^*$ (the optimal removal sequence) reveals that strokes are typically removed in a balanced way across the character. In other words, entire parts (e.g., only the right or bottom radical) are not erased at once, implying that preserving the global outline is more critical than retaining a single partial 
    component.
    
    \item \textbf{Short vs.\ long strokes:}
    Shorter strokes tend to be removed first, indicating that less frequent or more distinctive (often longer) strokes are vital for recognition. In some instances, however, even long strokes disappear early, suggesting that stroke length alone does not determine significance; rather, the overall combination of strokes matters.
\end{itemize}

Based on these findings, we confirm that a character’s global outline should remain intact while retaining the less frequent or more distinctive strokes in order to sustain legibility. This observation aligns with our stroke-length analysis, which suggests that rare strokes are often relatively long and contribute disproportionately to maintaining character identity. We also confirmed that when a large number of strokes are removed, distributing the removals across the entire shape is generally more effective than discarding a single localized part, leading to better legibility under heavy simplification.
\par

Despite its contributions, this approach has limitations.
Since a character with $N$ strokes has $2^N$ possible removal patterns, handling characters with up to 30 strokes becomes computationally challenging due to the exponential growth in complexity. Moreover, although neural network-based classifiers approximate human legibility, their decisions do not always align perfectly with human perception. Furthermore, our method relies on a discriminative model that classifies characters without capturing their underlying generative structure.
\par

For future work, incorporating generative models that learn the distribution of character variations could provide a more robust framework for simplification. Optimizing the search process to efficiently handle large stroke counts and validating our approach through human-centered evaluations will also be essential for practical applications. Finally, while this study focuses exclusively on stroke removal, multi-stroke characters frequently exhibit stroke merging.
We plan to extend our method to include both stroke removal and merging, enabling a more dynamic approach to character simplification.
\par

\begin{credits}
\subsubsection{\ackname} 
This work was supported by JSPS KAKENHI Grant Number JP24K22308, JP25H01149, JP23KJ1723, and JST CRONOS-JPMJCS24K4, JST ACT-X JPMJAX23CR.
\end{credits}
\newpage
%
%
%
\bibliographystyle{splncs04}
\bibliography{mybib_update}

\end{document}